\documentclass[conference]{IEEEtran}
\usepackage{cite}
\usepackage{amsmath,amssymb,amsfonts}
\usepackage{graphicx}
\usepackage{textcomp}
\usepackage{xcolor}
\usepackage{algorithm}
\usepackage{algorithmicx}
\usepackage{algpseudocode}
\usepackage{booktabs}
\usepackage{multicol}
\usepackage{multirow}

\usepackage{pifont}
\newcommand{\cmark}{\ding{51}}%
\newcommand{\xmark}{\ding{55}}%

\def\BibTeX{{\rm B\kern-.05em{\sc i\kern-.025em b}\kern-.08em
    T\kern-.1667em\lower.7ex\hbox{E}\kern-.125emX}}
\begin{document}

\title{Two-Stage Stance Labeling: User-Hashtag Heuristics with Graph Neural Networks\\
\thanks{Army Research Office}
}

\author{\IEEEauthorblockN{1\textsuperscript{st} Joshua Melton}
\IEEEauthorblockA{\textit{Department of Computer Science} \\
\textit{UNC-Charlotte}\\
Charlotte, USA \\
jmelto30@charlotte.edu}
\and
\IEEEauthorblockN{2\textsuperscript{nd} Shannon Reid}
\IEEEauthorblockA{\textit{Department of Criminal Justice and Criminology} \\
\textit{UNC-Charlotte}\\
Charlotte, USA \\
sreid33@charlotte.edu}
\and
\IEEEauthorblockN{3\textsuperscript{rd} Gabriel Terejanu}
\IEEEauthorblockA{\textit{Department of Computer Science} \\
\textit{UNC-Charlotte}\\
Charlotte, USA \\
gterejan@charlotte.edu}
\and
\IEEEauthorblockN{4\textsuperscript{th} Siddharth Krishnan}
\IEEEauthorblockA{\textit{Department of Computer Science} \\
\textit{UNC-Charlotte}\\
Charlotte, USA \\
skrishnan@charlotte.edu}
}

\maketitle

\begin{abstract}
The high volume and rapid evolution of content on social media present major challenges for studying the stance of social media users. In this work, we develop a two stage stance labeling method that utilizes the user-hashtag bipartite graph and the user-user interaction graph. In the first stage, a simple and efficient heuristic for stance labeling uses the user-hashtag bipartite graph to iteratively update the stance association of user and hashtag nodes via a label propagation mechanism. This set of soft labels is then integrated with the user-user interaction graph to train a graph neural network (GNN) model using semi-supervised learning. We evaluate this method on two large-scale datasets containing tweets related to climate change from June 2021 to June 2022 and gun control from January 2022 to January 2023. Our experiments demonstrate that enriching text-based embeddings of users with network information from the user interaction graph using our semi-supervised GNN method outperforms both classifiers trained on user textual embeddings and zero-shot classification using LLMs such as GPT4. We discuss the need for integrating nuanced understanding from social science with the scalability of computational methods to better understand how polarization on social media occurs for divisive issues such as climate change and gun control.
\end{abstract}


\section{Introduction}
\label{sec:introduction}

In today's digital society, online social media have become the main platform for dissemination of partisan and political information as well as the main forum for public discussion of political and social issues. The immense scale and interconnected nature of social media has meant that the scope of public discussion online permits varied and diverse viewpoints about a broad range of topics. Despite the potential for the co-existence of diverse perspectives online, research has shown that online communities are increasingly \textit{polarized}, especially with respect to contentious issues~\cite{climate-twitter,polarization-translation,social-media,quantify}. The increasingly polarized nature of online communities poses a real challenge to many democracies in today's world. Studies have shown that over the last 30 years, both Democrats and Republicans have become more negative in their views towards the opposition party~\cite{quantify}, and these negative views have affected outcomes in areas such as scholarship fund allocation, mate selection, employment decisions, and acceptance of public health measures~\cite{jiang2021social,peretti2020future}. Over the last several years, social media platforms like Twitter, Reddit, YouTube, and others have amassed millions of users who engage both with mainstream and fringe media outlets as well as engaging directly with other users on numerous political and social issues. As such, large social media platforms like Twitter present an excellent opportunity to study web-scale user behavior and polarization among differing viewpoints on contentious issues such as climate change and gun control.

The study of polarization online has broadly been divided into two main categories. The majority of computational social science research has focused on analyzing \textit{interactional polarization}, which occurs when users interact to a large extent exclusively with like-minded individuals in highly balkanized communities and are only minimally exposed to users with opposing viewpoint. Such \textit{echo chambers} isolate users in an in-group community that reinforces users' existing beliefs and skews user perspectives towards a particular issue~\cite{echochamber}. Another important form of polarization that is less well-studied is known as \textit{affective polarization}. Affective polarization refers to highly negative sentiments expressed by users towards out-groups of users with opposing viewpoints~\cite{social-media,climate-twitter}. Prior work on affective polarization has been limited in part due to reliance on manually annotated datasets and challenges in identifying user stances and group memberships in a computationally efficient manner. With recent advances in language modeling, stance detection, sentiment analysis, and network science-based approaches to community detection, researchers have begun to study affective polarization using large, unannotated corpora and interaction networks. Despite this recent progress, much work on affective polarization remains \textit{ad hoc} and incommensurate between studies.

Stance labeling remains a particularly challenging task due to the informal and noisy nature of social media language, the presence of sarcasm and irony, and the potential biases introduced by human annotators and automated models. To further facilitate research into polarization online, we develop a stance labeling method that incorporates both textual content and user social interactions in a scalable two-stage pipeline. In the first stage, we apply a reciprocal label propagation algorithm to the user-hashtag bipartite graph that generates a soft-labeled set of users. In the second stage, we construct a signed, weighted, and attributed user-user interaction network and use transformer language models to represent the content of user posts and to determine the sentiment of interactions between users. Using the soft labels generated from the label propagation stage, we employ semi-supervised training to train a graph neural network classifier to classify user stances.

Our stance labeling approach utilizes an understanding about linguistic and network homophily. Linguistic homophily states that people who share similar ideologies also share similar textual content, not only through similar keyword and hashtag usages but also through sentiment and linguistic features~\cite{yang2017overcoming,kovacs2020simil}, and the inverse phenomenon---that users with dissimilar views use different language---has also been investigated by researchers~\cite{polarization-translation}. Likewise, the concept of network homophily, which describes the phenomenon that similar nodes tend to be connected to one another has long been studied in graph theory~\cite{mcpherson2001birds,kossinets2009origins} and informs the success of graph neural networks~\cite{kipf-gcn, graphsage, ginsum}. By combining the content-based and network-based approaches into a single, scalable pipeline for stance labeling, we facilitate further large-scale analysis of both interactional and affective polarization online.

In sum, the contributions of this work are:
\begin{itemize}
    \item We develop an approach for estimating user stance labels that incorporates both textual features and social network interactions.
    \item A two-stage pipeline allows our approach to scale to large datasets with soft labeling and semi-supervised training.
    \item We discuss challenges and potential biases in stance determination of users when relying on short text content as commonly found on social media.
\end{itemize}

\section{Related Work}

As social media has become the primary forum for discussion on many topics, including political issues, there has been growing interest in user stance detection~\cite{aldayel2021survey,darwish2020cluster,jiang2023retweet}. Many existing methods can be divided into two broad categories: textual-based methods that use the content of messages, and network-based methods that use social or conversational structure. Textual-based methods use the content of tweets, user profiles, and hashtags to determine a user's stance on a particular topic. Several studies have used supervised classification of user stance labels where classifiers are trained using a set of features encoding tweet text, hashtags, and user profile information. These techniques have been applied to determine user political ideology~\cite{addawood2019linguistic,badawy2018russia}. These approaches all generally rely on manually annotated sets of users to train the classifier models, and such methods are often limited by the size and diversity of the training content as manual annotation is both time-consuming and expensive. Furthermore, these methods can be susceptible to annotator bias and poor generalizability due to the limited distribution of training examples.

Topic modeling and user clustering are two unsupervised techniques that have been employed for stance labeling. These unsupervised methods generally embed text or user profile information in a latent feature space and then apply a variety of topic modeling or cluster algorithms to identify texts of similar topics or clusters of similar users. While such techniques avoid the issues with manual annotation for supervised training, they suffer from the ~\textit{curse of dimensionality}---the search space for a solution grows exponentially with the increase in dimensionality---as there are many more possible patterns than in lower-dimensional subspaces. Selecting useful and informative features, and the computation and memory complexity of clustering methods on high-dimensional create problems for unsupervised clustering-based methods. For high-dimensional data, many clustering techniques fail to produce meaningful clusters, but on the other hand, such techniques can be very efficient on low-dimensional features spaces~\cite{darwish2020cluster}. Several studies have therefore employed dimensionality reduction techniques prior to clustering to first project content embeddings to a low-dimensional space, and such techniques can be robust to class imbalance between different stance groups.

Network-based approaches use the structure of social interaction networks to determine user ideology or stance. Such methods employ techniques such as label propagation, min-cut, and node embedding techniques to leverage structural information from retweet~\cite{jiang2023retweet} and user follower~\cite{barbera2015twitter} networks. Xiao et al.~\cite{xiao2020timme} generate a multi-relational network using retweets, mentions, likes, and follows for binary classification of user ideologies.

Recent works have combined both textual content and network features in socially-infused text mining. Li and Goldwasser~\cite{li2019encoding} combine user interactions and user sharing of news media to detect bias in news articles, and Johnson, Jin, and Goldwasser~\cite{johnson2017modeling} combine multiple representations from lexical features and social interactions to categorize tweets by politicians on a variety of political topics. Pan et al.~\cite{pan2016tri} use the network structure and node content to learn node representations for classifying the category of scientific publications. Such works combining textual and network information have demonstrated improvements in user stance deterction compared with methods that rely solely on textual content or network structure alone.

Our approach is likewise informed by socially-infused text mining that combines information from text and network relationships for user stance labeling, but the approach we present in this paper is unique from the methods described above in that we i) combine language features from state-of-the-art transformer language models with the structure and sentiment of social interactions using graph neural networks, and ii) our two-stage process allows our method to scale to large datasets with minimal supervision.

\section{Methods}

This section describes our proposed method to estimate the stance of Twitter users on political issues. We first describe the collection of data for the gun control and climate changes datasets. Then we characterize the hashtag-based heuristic method to generate soft labels for two stance groups of users, which are used as seed users for training stance labeling methods. We then introduce several baseline methods followed by our GNN-based approach for stance labeling.

\subsection{Data Collection}

We collect tweets related to two prominent and contentious political issues---climate change and gun control. The keywords listed in Table~\ref{tab:keywords} are used to scrape the initial set of relevant tweets related to each topic. To fully capture conversation cascades and user interactions, we then recursively collect all referenced tweets from each tweet in the initially matched set. This ensures that all available tweets from relevant conversations are included in our analysis, even if particular tweets do not explicitly use any of the keywords. The climate change dataset consists of tweets published between June 1st, 2021 and May 31st, 2022. In all, the climate change dataset consists of a total of 46M tweets authored by 4.8M unique users and contains 726,378 conversation threads of at least three tweets. The gun control dataset consists of tweets published between January 1st, 2022 and December 31st, 2022. In all, the gun control dataset consists of a total of 14.4M tweets from 2.66M unique users containing 335,000 conversation threads of at least three tweets.

\begin{center}
\begin{table}
\centering
\caption{Set of keywords used to collect tweets related to Climate Change and Gun Control from Twitter.}
    \begin{tabular}{ |c|c| } 
     \hline
     Climate Change & Gun Control \\
     \hline
      climate change & gun control \\
      global warming & gun rights \\
      climate hoax & second amendment \\
      global warming hoax & 2nd amendment \\
      global cooling & \#guncontrol \\
      \#ActOnClimate & \#guncontrolnow \\
      \#ClimateChange & \#gunreform \\
      \#climatechangehoax & \#gunviolence \\
      \#globalwarminghoax & \#endgunviolence \\
      \#globalcooling & \#2a \\
      \#globalwarmingisahoax & \#nra \\
      \#climatehoax & \#gunrights \\
       & \#secondamendement \\
       & \#shallnotbeinfringed \\
       & \#righttobeararms \\
     \hline
    \end{tabular}
    \label{tab:keywords}
\end{table}
\end{center}

\begin{figure}[h]
    \begin{center}
        \includegraphics[width=8cm]{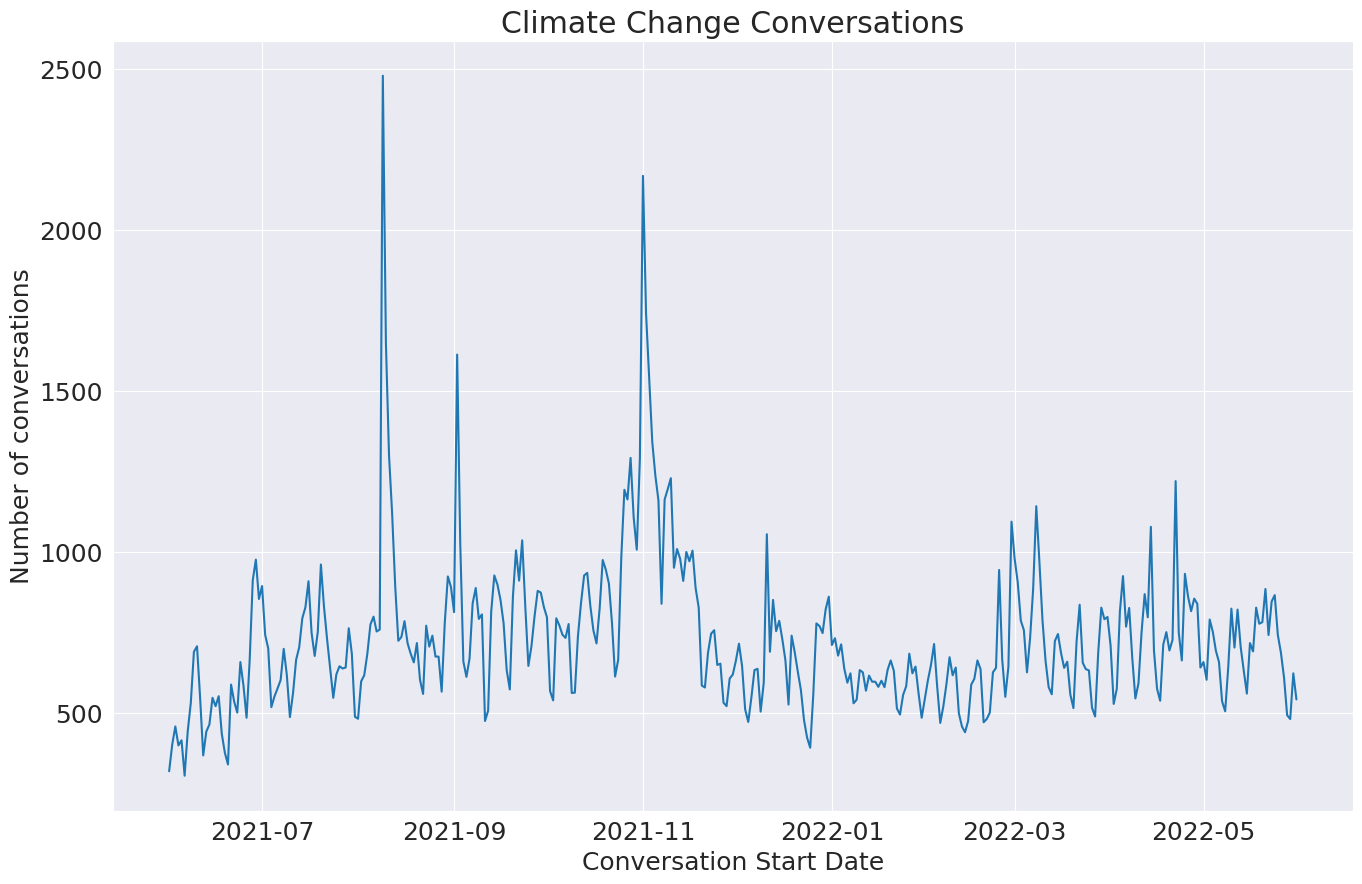}
    \end{center}
    \caption{Timeline of the number of Twitter conversations about climate change started each day.}
    \label{fig:climate-conv-timeline}
\end{figure}

\begin{figure}[h]
    \begin{center}
        \includegraphics[width=8cm]{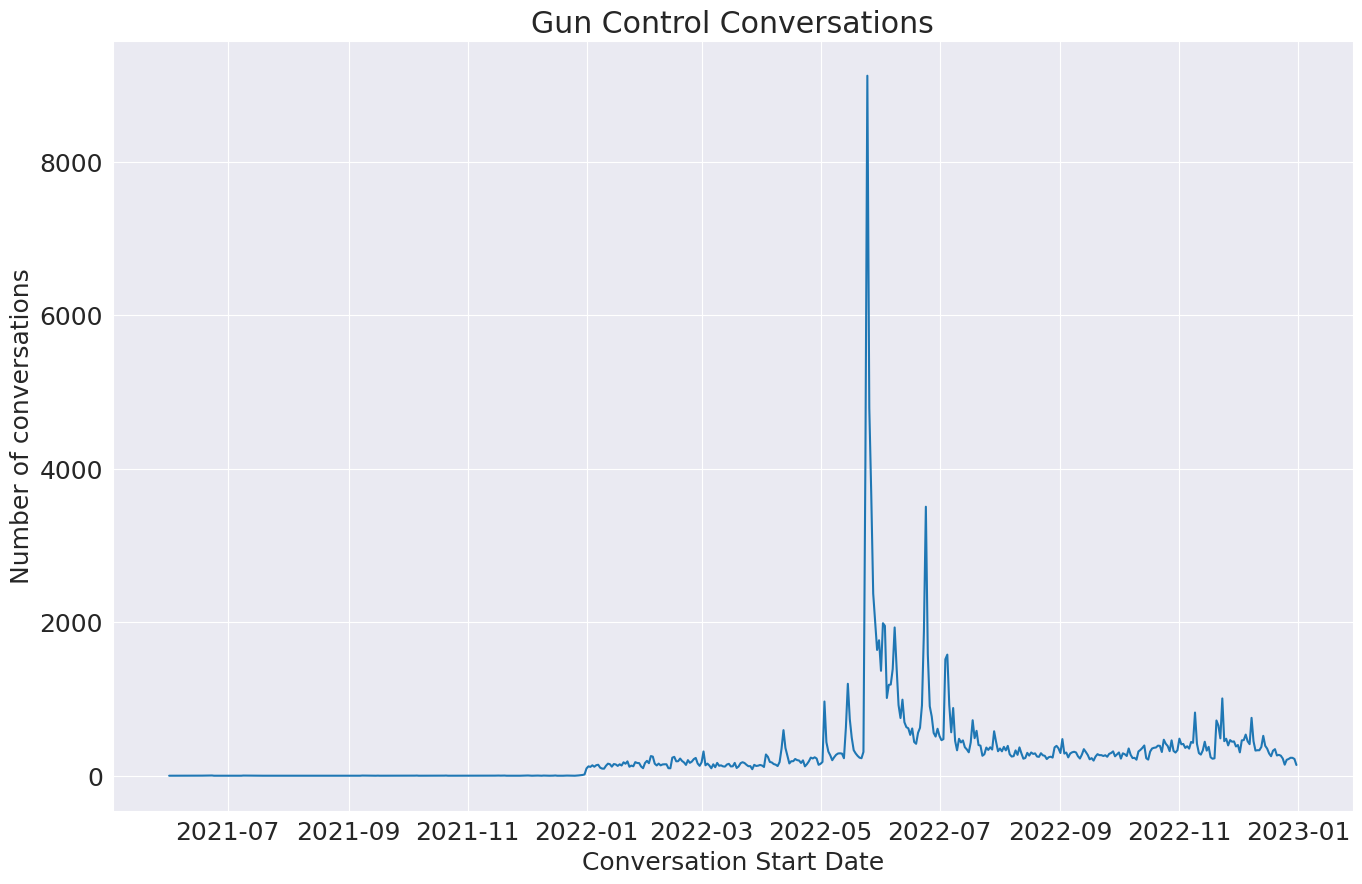}
    \end{center}
    \caption{Timeline of the number of Twitter conversations about gun control started each day.}
    \label{fig:gun-conv-timeline}
\end{figure}

These issues are good models for understanding the differences in behavior on social media for topics that are highly event driven, like gun control, and those that are less focused on particular events, like climate change. Figure~\ref{fig:climate-conv-timeline} illustrates that conversation threads related to climate change are started at a fairly consistent rate over time with some spikes in conversation activity connected with specific events. But compared with the conversation activity on Twitter surrounding gun control, illustrated in Figure~\ref{fig:gun-conv-timeline}, the frequency of discussions surrounding gun control spikes sharply in response to mass shooting events. The summer of 2022 saw several deadly and highly publicized mass shootings in Buffalo, Uvalde, and Chicago that generated large amounts of discourse surrounding gun control and gun violence on social media, illustrated by the sharp spikes from the end of May through late July. Shortly after mass shooting events though, the level of discourse abates, returning to near baseline levels between early August until November, when a shooting event at UVA caused a small spike in conversation activity. Event driven issues such as gun control exhibit different posting behavior and user interactions that we hypothesize lead to differences in polarization and require different techniques for intervention and mitigation of polarization on social media. Developing approaches for stance labeling that are both topic independent and robust to different types of discourse online is an important step for facilitating further analysis of polarization online.

\subsection{User-Hashtag Stance Labeling}

From the set of collected tweets for a given topic, we first construct a bipartite graph connecting users with the hashtags that each user posted. The user-hashtag bipartite graph $\mathcal{G}=(V, E)$ consists of two disjoint node sets containing the users and the hashtags posted in tweets by the user set. An edge $(u, h) \in E$ indicates that a user $u$ posted a tweet containing hashtag $h$. The weight $w(u, h)$ represents the number of times the author posted a particular hashtag across all of their authored tweets. Using the bipartite user-hashtag graph, we then develop a reciprocal label propagation-based method for automatically assigning stance labels to users and hashtags.

\begin{center}
\begin{table}
\centering
\caption{Sets of seed hashtags used for heuristic stance labeling of users in the climate change and gun control datasets.}
    \begin{tabular}{ |c|c| } 
     \hline
     \multicolumn{2}{|c|}{Climate Change} \\
     \hline
     Believe & Disbelieve \\
     \hline
        actonclimate & climatechangehoax \\
        climatecrisis & globalwarminghoax \\
        climateaction & globalcooling \\
        climateemergency & globalwarmingisahoax \\
        climateactionnow & climatehoax \\
     \hline
     \hline
     \multicolumn{2}{|c|}{Gun Control} \\
     \hline
     Pro & Anti \\
     \hline
        guncontrolnow & shallnotbeinfringed \\
        endgunviolence & righttobeararms \\
        gunreform & gunrights \\
     \hline
    \end{tabular}
    \label{tab:seeds}
\end{table}
\end{center}

 Based on a small (3-5) set of seed hashtags associated with each stance, we first propagate the stance associated with each seed hashtag across the user-hashtag bipartite graph to the set of users who posted a given labeled hashtag. The importance of a hashtag's stance to a particular destination user is weighted by the number of times the user posted each hashtag, as described in Algorithm~\ref{alg:propHtoU}. Users are then assigned a label based on the number of hashtag usages for each stance group. User labels are then propagated back to the hashtags used by each user, as described in Algorithm~\ref{alg:propUtoH}. Each hashtag is then scored based on the difference between the normalized count of usages from users of each stance group. Hashtag scores are normalized via min-max normalization, and the mean and standard deviation of hashtag scores is computed. Hashtags that score one or more standard deviations above or below the mean are assigned the corresponding stance label. We then repeat alternating label propagation until the model converges, i.e. the set of labeled users is the same from the prior iteration to the next.

\begin{algorithm}
\caption{Bipartite stance label propagation algorithm}\label{alg:bilabel}
    \begin{algorithmic}[1]
    \Require Bipartite user-hashtag graph $G$, seed hashtag set $H^{0}$
    \State $H^{0} \gets \{(h, s) \in {\textrm{Set of labeled seed hashtags}}\}$
    \State $U^{0} \gets \emptyset$
    \While {$U^{(k-1)} \neq U^{k}$}
        \State {$U^{k} \gets propagateTagsToUsers \left(G, \, H^{(k-1)} \right)$}
        \State {$H^{k} \gets propagateUsersToTags \left(G, \, U^{k} \right)$}
    \EndWhile
    \State \Return {$U^{K}, \, H^{K}$}
    \end{algorithmic}
\end{algorithm}

\begin{algorithm}
\caption{Propagate hashtag labels to users}\label{alg:propHtoU}
    \begin{algorithmic}[1]
    \Require Bipartite user-hashtag graph $G$, labeled hashtag set $H$
    \State {\textrm{user\_counts} = \{\}}
    \For {$h, s \in H$}
        \For {$u, w \in \mathcal{N}(h)$}
            \State{$\textrm{user\_counts}[u][s] \mathrel{+}= w$}
        \EndFor
    \EndFor
    \For {$u \in \textrm{user\_counts}$}
        \State {$U \gets (u, \, \max_{s}\left( \textrm{user\_counts[u]}\right))$}
    \EndFor
    \State \Return {$U$}
    \end{algorithmic}
\end{algorithm}

\begin{algorithm}
\caption{Propagate user labels to hashtags}\label{alg:propUtoH}
    \begin{algorithmic}[1]
    \Require Bipartite user-hashtag graph $G$, labeled user set $U$
    \State {\textrm{tag\_counts} = \{\}}
    \For {$u, s \in U$}
        \For {$h \in \mathcal{N}(u)$}
            \State {$\textrm{tag\_counts}[h][s] \mathrel{+}= 1$}
        \EndFor
    \EndFor
    \For {$h \in \textrm{tag\_counts}$}
        \State {$h_{score} = \frac{\textrm{tag\_counts}[h][s_2]}{|U_{s2}|} - \frac{\textrm{tag\_counts}[h][s_1]}{|U_{s1}|}$}
    \EndFor

    \State {$\mu = \textrm{mean}(h_{score} \, | \, h \in \textrm{tag\_counts})$}
    \State {$\sigma = \textrm{stdev}(h_{score} \, | \, h \in \textrm{tag\_counts})$}

    \For {$h \in \textrm{tag\_counts}$}
        \If {$h_{score} \geq \mu + \sigma$}
            \State {$H \gets (h, s_1)$}
        \ElsIf {$h_{score} \leq \mu - \sigma$}
            \State {$H \gets (h, s_2)$}
        \EndIf
    \EndFor
    \State \Return {$H$}
    \end{algorithmic}
\end{algorithm}

While the user-hashtag bipartite label propagation method can quickly label a number of users with high precision. It suffers from the limitation that a large proportion of users do not use hashtags, and while it is well suited for determining hashtags that are highly associated with one particular stance group as compared with the other, hashtags that are used across both stances albeit in different contexts will remain unlabeled.


\subsection{GNN Stance Labeling}

The user-hashtag stance labeling method can efficiently determine the stance group of users and discover important hashtags associated with each stance group, but it is limited by the fact that not all users post hashtags and that hashtags consist of only a small part of the overall textual content contained in tweets. Furthermore, the method does not take into account any social interactions between users. The hashtag-based heuristic has high precision but relatively low recall due to the limited breadth of users that stance labeling method covers. To develop a scalable model capable of labeling users in large-scale datasets, we employ the user-hashtag bipartite method as a means to generate a soft-labeled set of users from which we train a graph neural network classifier over the attributed user-user interaction graph.

We define a user interaction as a retweet, mention, reply, or quote from the author of the tweet towards another user. We then construct a weighted, signed, and attributed user-user interaction network $G = (V, E, X)$, where $V$ is the set of users and a directed edge $(u, v)_i \in E$ connects the author user to the target user of their interaction as defined above. The edge weight $w(u, v)_i$ corresponds to the sentiment of the tweet, scored by a BERTweet-based sentiment analysis model. BERTweet is a RoBERTa model fine-tuned on English tweets and trained on the SemEval 2017 corpus~\cite{nguyen-etal-2020-bertweet}. To characterize the overall interactions between users, we consolidate the multiple edges between pairs of users corresponding to individual tweets by combining each directed edge $(u, v)_i$ into a single edge for each user pair $(u, v)$ with the composite edge weight $w(u, v)$ computed as the mean sentiment score across all tweets between two users. Each user $u \in V$ is also associated with a feature vector $x_u \in X$ which corresponds to the element-wise mean of the BERTweet embeddings of all tweets authored by the user. We train a graph neural network model as a stance label classifier using semi-supervised learning with the graph partially labeled with the set of seed users labeled by our hashtag-based heuristic as a training set.

For the GNN, we compare two implementations from the prominent families of graph neural networks---convolutional and attentional. For convolutional GNNs, the neighbor coefficients are fixed weights; whereas, in attentional GNNs these weights are computed implicitly by an attention mechanism. Specifically, we evaluate graphSAGE\cite{graphsage} as an example of a convolutional GNN and graph attention network (GAT)~\cite{gat} as an example attentional GNN.

GraphSAGE defines a message passing function as:

\begin{equation}
    \boldsymbol{\hat{h}}_{v}^{(k)} = \sigma \left(\boldsymbol{\mathrm{W}} \sum_{u \in \mathcal{\tilde{N}}(v)} \frac{1}{|\mathcal{\tilde{N}}|} \boldsymbol{h}_{u}^{(k-1)} \right)
\end{equation}

\noindent
where $\mathcal{\tilde{N}}(v)$ is the local neighborhood of node $v$ with added self-loop, $\boldsymbol{\mathrm{W}}$ is a trainable weight matrix, and $\sigma$ is a non-linear activation function. We use the mean aggregator to aggregate the neighboring node features.

\noindent
We also consider a GAT layer operator, defined as:

\begin{equation}
    \boldsymbol{\hat{h}}_{v}^{(k)} = \sigma \left(\boldsymbol{\mathrm{W}} \sum_{u \in \mathcal{\tilde{N}}(v)} a_{vu} \boldsymbol{h}_{u}^{(k-1)} \right)
\end{equation}

\noindent
where aggregation coefficient $a_{vu}$ is computed dynamically in a node-dependent manner via edge softmax attention. In this case, the GAT function is a first-order approximation of a convolution applied over the weighted adjacency matrix, where the edge weights are computed via the attention mechanism.

\subsection{Baselines}

\begin{table}
    \centering
    \footnotesize
    
    {
    \caption{Models utilized for user stance labeling and indication whether the models incorporate textual based information and/or network structural information.}
    \begin{tabular}{llcc}
        \toprule 
        Model Type & Model  & Text Content & Network Structure \\
        \midrule
        {\textit{Random}} & Weighted random & \xmark & \xmark  \\
        \midrule 
    \multirow{4}{*}{\textit{Transformer}} 
        & BERT-base-uncased  & \cmark & \xmark \\
        & BERT-large-uncased & \cmark & \xmark \\
        & RoBERTa-base & \cmark & \xmark \\
        & RoBERTa-large & \cmark & \xmark \\
        & BERTweet  & \cmark & \xmark \\
        & GPT-4 & \cmark & \xmark \\
    \midrule
    \multirow{3}{*}{\textit{GNN}} 
        & GraphSAGE & \cmark & \cmark \\
        & GAT & \cmark & \cmark \\ 
    \bottomrule
    \end{tabular}
    }
    \label{tab:models}
\end{table}

We compare our stance labeling method against six transformer language models that leverage the textual content of tweets to represent users. With each language model, we compute a user embedding by computing the element-wise mean of the embedding of all the users' tweets from BERT, RoBERTa, and BERTweet models. Using the set of labeled users from our heuristic method, we train an MLP classifier to predict user stance label.

We evaluate classification performance of the stance labeling methods using a set of manually annotated users and a set of users annotated via GPT-4 using zero-shot classification. Two social science experts independently annotated a set of 250 users for the climate change dataset and 350 users for the gun control dataset by analyzing the top 20 most popular tweets from each user. We provide the same set of tweets to GPT-4 and prompt the LLM to provide the sentiment of content with respect to gun control or climate change. For the gun control dataset, the function call prompt is: ``The stance of the content. Is the message pro gun control, anti gun control, or neutral?'' with three classification options of ``pro'', ``anti'', or ``neutral''. For the climate change dataset, ``The stance of the content. Does the message capture the user's belief in climate change, disbelief, or is it neutral?'' with classification options of ``belief'', ``disbelief'', or ``neutral''. We then aggregate the stance labeling results for each individual tweet to determine a user's overall stance on the issue of gun control or climate change. Users who had a majority of tweets classified as belonging to one stance were labeling accordingly and those users whose tweets were labeled as all neutral are classified as undetermined. In the case where there is an equivalent number of tweets labeled as both stances, the user is labeled as undetermined.

Using the set of manual and GPT-4 annotated users for the two datasets, we compare the classification performance of both text-based methods and our GNN-based method informed by socially infused text mining. We additionally compare the performance of GPT-4 for zero-shot stance labeling of users compared against the manual annotation by domain experts.

\section{Results}

We report the macro averaged precision, recall, and F1 score for each model, and we use the F1 score as the primary metric due to the class imbalance in the datasets. Metrics reported are averaged across five trials for all models.

\subsection{Gun Control Dataset}

\begin{table*}[t]
    \centering
    \footnotesize
    {
    \caption{Results for stance labeling classification on the \texttt{Gun Control} dataset ($N=350$). The best F1 (macro) scores for each model type are shown in bold and the best overall scores are underlined.
    }
    \begin{tabular}{l ccc ccc}
        \toprule 
        & \multicolumn{3}{c}{\texttt{GPT4 Labels}} & \multicolumn{3}{c}{\texttt{Manual Labels}}\\
        \cmidrule(lr){2-4} \cmidrule(lr){5-7}
        Model & Prec. & Recall & F1 & Prec. & Recall & F1 \\
        \midrule
        Weighted random & 47.36 & 49.99 & 39.45 & 40.76 & 49.92 & 38.11 \\
        \midrule 
        BERT-base-uncased & 84.25 & 85.38 & 84.66 & 83.08 & 82.75 & 82.88 \\
        BERT-large-uncased & 80.47 & 82.16 & 79.96 & 81.44 & 81.71 & 81.13 \\
        RoBERTa-base & 83.33 & 83.94 & 83.60 & 84.79 & 83.99 & 84.25 \\
        RoBERTa-large & 81.39 & 82.69 & 79.52 & 84.41 & 84.18 & 83.14 \\
        BERTweet & 83.59 & 85.43 & 82.97 & 85.45 & 85.76 & 85.13  \\
        GPT-4 & - & - & - & 89.81 & 85.17 & 87.05 \\
    \midrule 
    \midrule
        GraphSAGE & 87.58 & 88.75 & 88.02 & 91.09 & 90.46 & 90.70 \\
        GAT & 87.17 & 88.99 & 87.56 & 91.28 & 91.46 & 91.36 \\ 
    \bottomrule
    \end{tabular}
    }
    \label{tab:gun-results}
\end{table*}

The classification results for the stance labeling models on the gun control dataset are shown in Table~\ref{tab:gun-results}. On the gun control dataset, BERT and RoBERTa transformer methods achieve a macro averaged F1 score of between 82\%-84\% when evaluated on the set of manually annotated users. BERTweet, having been finetuned on tweets, surpasses the performance of the other transformer models achieving an F1 score of 85.13\%. GPT-4 zero-shot classification outperforms the trained BERTweet classifier by two percent on F1 score and a four percent increase in precision while maintaining an equivalent recall performance.

For the gun control dataset, the GNN-based models that incorporate network structure information with the content-based representations of users produced by BERTweet, outperform both the trained text-based classifiers and zero-shot classification with GPT-4. Using graphSAGE and GAT to enrich users' content-based representations results in an improvement in stance labeling classification compared with the unenriched BERTweet embeddings. With GAT, incorporating users' local neighborhood information with their text-based embeddings improves classification performance by four percent in F1 score. Compared with zero-shot classification using GPT-4, the BERTweet + GAT model performs 3\%-4\% better in F1 score on classifying user stance labels for the manually labeled user set. Compared with graphSAGE, which takes the mean representation of a node's neighborhood effectively weighting each incident edge on the node equally, GAT utilizes edge softmax attention to dynamically compute edge weights when aggregating node neighborhood information. The dynamic computation of edge importance for users results in a marginal improvement in classification performance.

\subsection{Climate Change Dataset}

\begin{table*}[t]
    \centering
    \footnotesize
    {
    \caption{Results for stance labeling classification on the \texttt{Climate Change} dataset ($N=250$). The best F1 (macro) scores for each model type are shown in bold and the best overall scores are underlined.
    }
    \begin{tabular}{l ccc ccc}
        \toprule 
        & \multicolumn{3}{c}{\texttt{GPT4 Labels}} & \multicolumn{3}{c}{\texttt{Manual Labels}}\\
        \cmidrule(lr){2-4} \cmidrule(lr){5-7}
        Model & Prec. & Recall & F1 & Prec. & Recall & F1 \\
        \midrule
        Weighted random & 47.36 & 49.99 & 39.45 & 45.26 & 47.55 & 42.87 \\
        \midrule 
        BERT-base-uncased & 76.74 & 79.82 & 77.36 & 78.60 & 82.54 & 79.36 \\
        BERT-large-uncased & 75.63 & 79.14 & 73.86 & 76.11 & 80.28 & 74.17 \\
        RoBERTa-base & 76.30 & 79.92 & 76.21 & 76.63 & 80.92 & 76.50 \\
        RoBERTa-large & 75.72 & 79.36 & 75.16 & 76.91 & 81.33 & 76.29 \\
        BERTweet & 77.26 & 80.78 & 77.64 & 78.35 & 82.66 & 78.78 \\
        GPT-4 & - & - & - & 85.36 & 86.08 & 85.71 \\
    \midrule 
    \midrule
        GraphSAGE & 84.62 & 86.06 & 85.25 & 83.80 & 85.91 & 84.67 \\
        GAT & 83.73 & 86.43 & 84.70 & 83.47 & 87.02 & 84.62 \\ 
    \bottomrule
    \end{tabular}
    }
    \label{tab:climate-results}
\end{table*}

The classification results for the stance labeling models on the climate change dataset are shown in Table~\ref{tab:climate-results}. Compared with the gun control dataset, where pro gun control users outnumber anti gun control users by roughly a ratio of 2:1 in the heuristic labeled user set, the imbalance in the climate change dataset is significant with climate change believers outnumbering disbelievers by a ratio of nearly 10:1. We also observe that climate change related conversations are less topic-focused than gun control discussions on Twitter. Due in part to the severe class imbalance in the dataset in addition to the less focused of climate change conversation activity on Twitter, stance determination of users in the climate change dataset is more difficult than in the gun control dataset. 

Of the BERT and RoBERTa based classifier models, BERT-base-uncased performs the best with an F1 score of 79.36\%, narrowly outperforming BERTweet with an F1 score of 78.78\%. Enriching the BERTweet user embeddings with structural information from the user interaction graph improves the classification performance by six percent to 84.67\% F1 score. Despite the performance uplift from the user social network, for the climate change dataset zero-shot classification with GPT-4 outperforms all other models with an F1 score of 85.71\%, outperforming other transformer methods by six percent and GNN-based approaches by one percent.

In all, enriching text-based representations of users with structural information from the user-interaction graph results in a significant improvement in classification performance compared with the text-based embeddings alone. Zero-shot classification of user stances using GPT-4 consistently outperforms BERT and RoBERTa-based classifier models, but incorporating social network information surpasses GPT-4 classification in the gun control dataset and significantly narrows the performance gap in the climate change dataset.

\section{Discussion}

When researching online polarization, there are serious divides in how questions regarding user stance are structured, and this divide is also reflected in how different researchers integrate the literature and findings into future work. Oftentimes, this split can be seen in how social science researchers integrate computer science research into their studies and vice versa. This methodological gap can inhibit the understanding of the phenomena of online polarization and radicalization. Big data research can provide scalability not feasible by traditional qualitative social science research, while social science expertise provides understanding of the subtleties of these interactions on human behavior. As has been seen in the aftermath of events like radicalized mass shooters, the online behavior of these individuals is not well understood and the need to understand how and why these processes take place online needs to be examined. The integration of subject-matter expertise with AI is necessary for the large-scale application of big data methodology into human-level polarization research and policy work, requiring a balance of the nuance provided by social science subject-matter experts with the speed and volume provided by computation experts. Qualitative research offers methodological tools like narrative analysis to analyze the cultural and historical contingency of the terms, beliefs and issues narrators address in their writings~\cite{rosenwald1992storied}. The small narratives created from tweets can provide insights into the range of ongoing societal arguments that create collective identity online~\cite{coupland2004constructing}. The wealth of this data adds critical information to understanding how people discuss their thoughts and opinions about polarizing topics like gun rights or climate change, and the ability to combine these two approaches can greatly benefit the understanding of online polarization.

One area of difficulty in automating stance detection on topics like gun control or climate change is the need to dichotomize beliefs that really exist on a spectrum. The extreme ends of these beliefs are straight-forward to categorize, but the majority of people exist in the space in-between. A better goal is to qualitatively code the tipping point as people's stance moves between stances on issues. Automated stance analysis can mislabel users because of difficulties in discerning written cues like tone, connotation, coded words, or pictures. Some of these coding errors are due to a purposeful misdirection of a tweet that is clarified by the picture. For example, user 2945287090 said ``I'm pro-choice…are you? https://t.co/WtYW6zohSB'' but the picture has them wearing a shirt that said pro-choice with an array of guns to ``choose'' from. Since the picture cannot be assessed by many stance labeling algorithms, these misdirections would be potentially miscoded. 
There are also users who frame their arguments in neutral language or the language of their opposition. For example, user 66811907 states ``Gun control advocates do not have a monopoly on outrage or sorrow. They do not have a monopoly on ideas for addressing gun violence. They do not have a monopoly wanting to fix the problem.'' The larger context of this user's tweets balance between being pro-Second Amendment and being critical to the responses to mass shooting events, such as Uvalde. As a whole, this user falls more strongly on the side of being anti-gun control but certainly has criticism of open market gun purchasing. The neutral framing can be difficult to for automated approaches to parse since the full narrative of the users' tweets need to be taken into account.

However, the coding errors are not only on the side of computational approaches. One of the key limitations to using only human-based qualitative coding is that the deep understanding of stances is highly time-consuming and delayed. Especially in online space, the posting moves far faster than researchers can code by hand. This issue means that oftentimes qualitative researchers are forced to either focus on a subset of tweets or a subset of users, and when examining a subset of tweets to establish stance, there are bound to be errors. The balance between efficiency and accuracy means that there will be miscoded information that could potentially be corrected by looking at all the available data. The ability to integrate big data and computational methods can be a force multiplier for qualitative researchers as a mixed methods approach greatly improves the scope and generalizability of work on online polarization.

As mentioned previously, these stances exist on a spectrum and since the researcher is making a qualitative decision of when a person's stance moves between two ends, the middle ground can often be coded either way. For example, is a person who owns guns for hunting but pushes for very strong hand gun control measures, pro-gun control or anti-gun control? Does wanting strong background checks or other restrictions inherently make someone pro/anti gun? Even in the context of a large body of tweets, it may not be fully clear how this stance should be coded and is not necessarily an error on either side but rather a decision reflecting this balance. One example is user 66811907 who has a large body of tweets that different coders could code differently ``This is your friendly reminder that New York already bans ``assault weapons'', that the shooter procured his NY-compliant ``non-assault weapon'' legally, and that the cops who stopped him showed up with their ``assault weapon'' version of the same gun to combat a civilian threat.'' and ``Rifles of any kind account for fewer annual homicides than hands and feet. Mass shooters don't need pistol grips, etc., to cause horrific carnage, but on the flip side those features can be quite useful for lawful owners using them in self-defense. That's why cops carry them, too.'' This example is just one that highlights that neither AI nor human coding is without issues, error, or disagreement and the reporting of inter-rater reliability could be a useful measure to add for both human annotators and automated approaches. 

\section{Conclusion}

In this work, we take a step towards a broader understanding of user stance labeling through socially infused text mining---incorporating both textual content and network interactions using language models and graph neural networks. Both gun control and climate change are issues that have a range of historical contexts, political connotations, and intersections with race and gender. It is in this space that subject-matter expertise can provide the contextual information to inform the work of computational approaches. Social science insights provide the ground truths necessary to properly contextualize computational approaches to complex human stances to ensure that the models are acknowledging the real-world context, minimizing errors, and maximizing the relevance of the generated outputs to policymakers and to the broader research community.

\section*{Acknowledgment}

The research was sponsored by the Army Research Office and was accomplished under Grant Number W911NF-22-1-0035. The views and conclusions contained in this document are those of the authors and should not be interpreted as representing the official policies, either expressed or implied, of the Army Research Office or the U.S. Government. The U.S. Government is authorized to reproduce and distribute reprints for Government purposes notwithstanding any copyright notation herein.

\bibliographystyle{./IEEEtran}
\bibliography{./IEEEabrv,./biblio}

\begin{thebibliography}{10}
\providecommand{\url}[1]{#1}
\csname url@samestyle\endcsname
\providecommand{\newblock}{\relax}
\providecommand{\bibinfo}[2]{#2}
\providecommand{\BIBentrySTDinterwordspacing}{\spaceskip=0pt\relax}
\providecommand{\BIBentryALTinterwordstretchfactor}{4}
\providecommand{\BIBentryALTinterwordspacing}{\spaceskip=\fontdimen2\font plus
\BIBentryALTinterwordstretchfactor\fontdimen3\font minus
  \fontdimen4\font\relax}
\providecommand{\BIBforeignlanguage}[2]{{%
\expandafter\ifx\csname l@#1\endcsname\relax
\typeout{** WARNING: IEEEtran.bst: No hyphenation pattern has been}%
\typeout{** loaded for the language `#1'. Using the pattern for}%
\typeout{** the default language instead.}%
\else
\language=\csname l@#1\endcsname
\fi
#2}}
\providecommand{\BIBdecl}{\relax}
\BIBdecl

\bibitem{climate-twitter}
A.~Tyagi, J.~Uyheng, and K.~M. Carley, ``Affective polarization in online
  climate change discourse on twitter,'' in \emph{ASONAM}, 2020, pp. 443--447.

\bibitem{polarization-translation}
A.~R.~KhudaBukhsh, R.~Sarkar, M.~S. Kamlet, and T.~Mitchell, ``We don't speak
  the same language: Interpreting polarization through machine translation,''
  \emph{AAAI}, vol.~35, no.~17, pp. 14\,893--14\,901, May 2021.

\bibitem{social-media}
M.~Yarchi, C.~Baden, and N.~Kligler-Vilenchik, ``Political polarization on the
  digital sphere: A cross-platform, over-time analysis of interactional,
  positional, and affective polarization on social media,'' \emph{Political
  Communication}, vol.~38, no. 1-2, pp. 98--139, 2021.

\bibitem{quantify}
K.~Garimella, G.~D.~F. Morales, A.~Gionis, and M.~Mathioudakis, ``Quantifying
  controversy in social media,'' \emph{CoRR}, vol. abs/1507.05224, 2015.

\bibitem{jiang2021social}
J.~Jiang, X.~Ren, and E.~Ferrara, ``Social media polarization and echo chambers
  in the context of covid-19: Case study,'' \emph{JMIRx med}, vol.~2, no.~3, p.
  e29570, 2021.

\bibitem{peretti2020future}
P.~Peretti-Watel, V.~Seror, S.~Cortaredona, O.~Launay, J.~Raude, P.~Verger,
  L.~Fressard, F.~Beck, S.~Legleye, O.~L'Haridon, D.~Léger, and J.~K. Ward,
  ``A future vaccination campaign against covid-19 at risk of vaccine hesitancy
  and politicisation,'' \emph{The Lancet Infectious Diseases}, vol.~20, no.~7,
  pp. 769--770, 2020.

\bibitem{echochamber}
R.~Karlsen, K.~Steen-Johnsen, D.~Wollebæk, and B.~Enjolras, ``Echo chamber and
  trench warfare dynamics in online debates,'' \emph{European Journal of
  Communication}, vol.~32, no.~3, pp. 257--273, 2017, pMID: 28690351.

\bibitem{yang2017overcoming}
Y.~Yang and J.~Eisenstein, ``Overcoming language variation in sentiment
  analysis with social attention,'' \emph{Trans. Assoc. for Comput. Linguist.},
  vol.~5, pp. 295--307, 2017.

\bibitem{kovacs2020simil}
B.~Kovacs and A.~M. Kleinbaum, ``Language-style similarity and social
  networks,'' \emph{Psychological Science}, vol.~31, no.~2, pp. 202--213, 2020.

\bibitem{mcpherson2001birds}
M.~McPherson, L.~Smith-Lovin, and J.~M. Cook, ``Birds of a feather: Homophily
  in social networks,'' \emph{Annual review of sociology}, vol.~27, no.~1, pp.
  415--444, 2001.

\bibitem{kossinets2009origins}
G.~Kossinets and D.~J. Watts, ``Origins of homophily in an evolving social
  network,'' \emph{American journal of sociology}, vol. 115, no.~2, pp.
  405--450, 2009.

\bibitem{kipf-gcn}
T.~N. Kipf and M.~Welling, ``Semi-{Supervised} {Classification} with {Graph}
  {Convolutional} {Networks},'' in \emph{ICLR}, 2017.

\bibitem{graphsage}
W.~Hamilton, Z.~Ying, and J.~Leskovec, ``Inductive representation learning on
  large graphs,'' in \emph{NeurIPS}, 2017, pp. 1024--1034.

\bibitem{ginsum}
K.~Xu, W.~Hu, J.~Leskovec, and S.~Jegelka, ``How powerful are graph neural
  networks?'' in \emph{ICLR}, 2019.

\bibitem{aldayel2021survey}
A.~ALDayel and W.~Magdy, ``Stance detection on social media: State of the art
  and trends,'' \emph{Information Processing and Management}, vol.~58, no.~4,
  p. 102597, 2021.

\bibitem{darwish2020cluster}
K.~Darwish, P.~Stefanov, M.~Aupetit, and P.~Nakov, ``Unsupervised user stance
  detection on twitter,'' \emph{Proceedings of the International AAAI
  Conference on Web and Social Media}, vol.~14, no.~1, pp. 141--152, May 2020.

\bibitem{jiang2023retweet}
J.~Jiang, X.~Ren, and E.~Ferrara, ``Retweet-bert: political leaning detection
  using language features and information diffusion on social networks,'' in
  \emph{Proceedings of the International AAAI Conference on Web and Social
  Media}, vol.~17, 2023, pp. 459--469.

\bibitem{addawood2019linguistic}
A.~Addawood, A.~Badawy, K.~Lerman, and E.~Ferrara, ``Linguistic cues to
  deception: Identifying political trolls on social media,'' in
  \emph{Proceedings of the international AAAI conference on web and social
  media}, vol.~13, 2019, pp. 15--25.

\bibitem{badawy2018russia}
A.~Badawy, E.~Ferrara, and K.~Lerman, ``Analyzing the digital traces of
  political manipulation: The 2016 russian interference twitter campaign,'' in
  \emph{ASONAM}, 2018, pp. 258--265.

\bibitem{barbera2015twitter}
P.~Barberá, J.~T. Jost, J.~Nagler, J.~A. Tucker, and R.~Bonneau, ``Tweeting
  from left to right: Is online political communication more than an echo
  chamber?'' \emph{Psychological Science}, vol.~26, no.~10, pp. 1531--1542,
  2015.

\bibitem{xiao2020timme}
Z.~Xiao, W.~Song, H.~Xu, Z.~Ren, and Y.~Sun, ``Timme: Twitter
  ideology-detection via multi-task multi-relational embedding,'' in
  \emph{SIGKDD}, 2020, pp. 2258--2268.

\bibitem{li2019encoding}
C.~Li and D.~Goldwasser, ``Encoding social information with graph convolutional
  networks forpolitical perspective detection in news media,'' in \emph{ACL},
  2019, pp. 2594--2604.

\bibitem{johnson2017modeling}
K.~Johnson, D.~Jin, and D.~Goldwasser, ``Modeling of political discourse
  framing on twitter,'' in \emph{Proceedings of the International AAAI
  Conference on Web and Social Media}, vol.~11, no.~1, 2017, pp. 556--559.

\bibitem{pan2016tri}
S.~Pan, J.~Wu, X.~Zhu, C.~Zhang, and Y.~Wang, ``Tri-party deep network
  representation,'' in \emph{IJCAI}.\hskip 1em plus 0.5em minus 0.4em\relax
  AAAI, 2016, pp. 1895--1901.

\bibitem{nguyen-etal-2020-bertweet}
D.~Q. Nguyen, T.~Vu, and A.~Tuan~Nguyen, ``{BERT}weet: A pre-trained language
  model for {E}nglish tweets,'' in \emph{EMNLP}, Q.~Liu and D.~Schlangen,
  Eds.\hskip 1em plus 0.5em minus 0.4em\relax Association for Computational
  Linguistics, Oct. 2020, pp. 9--14.

\bibitem{gat}
P.~Veličković, G.~Cucurull, A.~Casanova, A.~Romero, P.~Liò, and Y.~Bengio,
  ``Graph {Attention} {Networks},'' in \emph{ICLR}, 2018.

\bibitem{rosenwald1992storied}
G.~C. Rosenwald and R.~L. Ochberg, \emph{Storied lives: The cultural politics
  of self-understanding}.\hskip 1em plus 0.5em minus 0.4em\relax Yale
  University Press, 1992.

\bibitem{coupland2004constructing}
C.~Coupland and A.~D. Brown, ``Constructing organizational identities on the
  web: A case study of royal dutch/shell,'' \emph{Journal of management
  studies}, vol.~41, no.~8, pp. 1325--1347, 2004.

\end{thebibliography}

\end{document}